\documentclass[letterpaper]{article} % DO NOT CHANGE THIS
\usepackage{aaai2026}  % DO NOT CHANGE THIS
\usepackage{times}  % DO NOT CHANGE THIS
\usepackage{helvet}  % DO NOT CHANGE THIS
\usepackage{courier}  % DO NOT CHANGE THIS
\usepackage[hyphens]{url}  % DO NOT CHANGE THIS
\usepackage{graphicx} % DO NOT CHANGE THIS
\usepackage{natbib}  % DO NOT CHANGE THIS AND DO NOT ADD ANY OPTIONS TO IT
\usepackage{caption} % DO NOT CHANGE THIS AND DO NOT ADD ANY OPTIONS TO IT
\usepackage{algorithm}
\usepackage{algorithmic}
\usepackage{amsfonts}
\usepackage{amsmath}

\usepackage{multirow}
\usepackage{makecell}
\usepackage{booktabs}
\usepackage{xspace}  

\newcommand{\sys}{\texttt{DEPO}\xspace}

\newcommand{\llama}{Llama3.1-8B-Instruct}
\newcommand{\qwen}{Qwen2.5-7B-Instruct}
\newcommand{\bcllama}{Llama3.1-8B-BC}
\newcommand{\bcqwen}{Qwen2.5-7B-BC}

\usepackage{newfloat}
\usepackage{listings}
\DeclareCaptionStyle{ruled}{labelfont=normalfont,labelsep=colon,strut=off} % DO NOT CHANGE THIS
\floatstyle{ruled}
\newfloat{listing}{tb}{lst}{}
\floatname{listing}{Listing}
\title{\sys: Dual‑Efficiency Preference Optimization for LLM Agents}

\author {
    Sirui Chen\textsuperscript{\rm 1,\rm 2,\rm 3\equalcontrib},
    Mengshi Zhao\textsuperscript{\rm 4\equalcontrib},
    Lei Xu\textsuperscript{\rm 2,\rm 5},
    Yuying Zhao\textsuperscript{\rm 3},\\
    Beier Zhu\textsuperscript{\rm 3}\thanks{Corresponding author.},
    Hanwang Zhang\textsuperscript{\rm 3},
    Shengjie Zhao\textsuperscript{\rm 1}\footnotemark[2],
    Chaochao Lu\textsuperscript{\rm 2}\footnotemark[2]
}
\affiliations {
    \textsuperscript{\rm 1}Tongji University,
    \textsuperscript{\rm 2}Shanghai Artificial Intelligence Laboratory,
    \textsuperscript{\rm 3}Nanyang Technological University\\
    \textsuperscript{\rm 4}The University of Hong Kong,
    \textsuperscript{\rm 5}École Polytechnique Fédérale de Lausanne (EPFL)\\
    2111292@tongji.edu.cn, \{beier.zhu, hanwangzhang\}@ntu.edu.sg, luchaochao@pjlab.org.cn
}

\usepackage{bibentry}
\begin{document}

\maketitle

\begin{abstract}
Recent advances in large language models (LLMs) have greatly improved their reasoning and decision-making abilities when deployed as agents. Richer reasoning, however, often comes at the cost of longer chain of thought (CoT), hampering interaction efficiency in real-world scenarios. Nevertheless, there still lacks systematic definition of LLM agent efficiency, hindering targeted improvements. 
To this end, we introduce dual‑efficiency, comprising 
(i) step-level efficiency, which minimizes tokens per step, and (ii) trajectory-level efficiency, which minimizes the number of steps to complete a task. 
Building on this definition, we propose \sys, a dual-efficiency preference optimization method that jointly rewards succinct responses and fewer action steps. Experiments on WebShop and BabyAI show that \sys cuts token usage by up to 60.9\% and steps by up to 26.9\%, while achieving up to a 29.3\% improvement in performance. \sys also generalizes to three out-of-domain math benchmarks and retains its efficiency gains when trained on only 25\% of the data. 
\end{abstract}

\begin{links}
    \link{Project page}{https://opencausalab.github.io/DEPO}
\end{links}

% ======================= [ SECTION: Introduction ] ===================== %

\section{Introduction}
\label{sec:introduction}

\begin{figure}[t!]  
\centering
\includegraphics[width=\columnwidth]{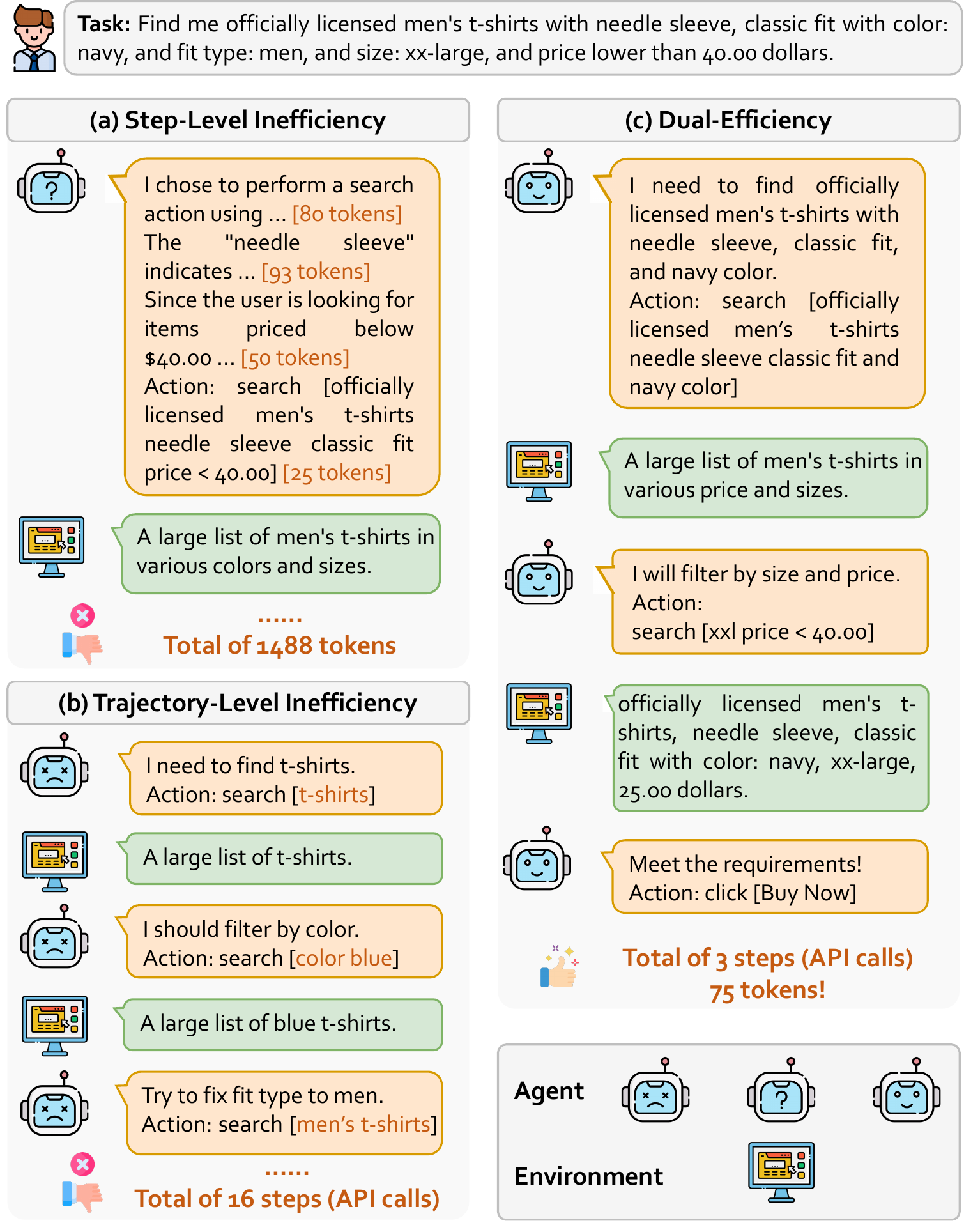}
  \caption{A comparison between (a) step-level inefficiency, arising from latency and cost in LLM token generation; (b) trajectory-level inefficiency, arising from latency and cost in environment interactions such as API calls, and our defined (c) dual-efficiency. For LLM agents, achieving genuine efficiency requires joint optimization across both dimensions.}
\label{fig:inefficiency}
\end{figure}

LLMs have shown impressive capabilities in complex tasks like long-context reasoning \citep{bai-etal-2024-longbench}, coding \citep{zhuo2025bigcodebench}, and mathematics \citep{luo2025wizardmath}. To bring these abilities closer to real-world applications, recent studies have started investigating the LLM agents, aiming to build autonomous systems capable of task planning, tool use, and multi-turn interaction with dynamic environments \citep{shi-etal-2024-direct,song-etal-2024-trial,chen-etal-2024-agent,zeng-etal-2024-agenttuning,yuan2025agent,fu2025agentrefine,wang2025ragen,feng2025group}. 
As advanced models pursue greater reasoning accuracy, their CoT \citep{wei2022chain,zhao2025unsupervised} often becomes longer, which leads to increased response latency and diminished execution efficiency.
Efficiency is crucial for transitioning LLM agents from prototypes to practical applications, as slow response times are a major barrier to real-world adoption. However, speed cannot come at the expense of accuracy—a fast but unreliable agent is useless, or even harmful. We therefore seek response efficiency while preserving reasoning quality. 

Current discussions on LLM efficiency predominantly focus on reducing the number of tokens generated~\citep{wang2025harnessing,qu2025survey,feng2025efficient}. 
However, this perspective overlooks the interaction dynamics of LLM agents. In practice, agents often need to make API calls or access web services at each decision step,  where latency and cost scale with the number of steps, not just tokens.
Therefore, we first define LLM agent efficiency along two key dimensions, which we term \emph{dual-efficiency}. 
(1) \emph{Step-level efficiency}: Aims to minimize the number of tokens generated in each interaction step, promoting concise responses.
(2) \emph{Trajectory-level efficiency}: Aims to minimize the total number of interaction steps required to complete a task. 
Figure \ref{fig:inefficiency} presents examples comparing LLM agents with step-level and trajectory-level inefficiencies, alongside agents achieving dual-efficiency. A step-level inefficient agent generates too many tokens per step due to overthinking (Figure \ref{fig:inefficiency}(a)). A trajectory-level inefficient agent is concise at each step but requires numerous steps to complete the task because of inaccurate reasoning (Figure \ref{fig:inefficiency}(b)). In contrast, a dual-efficient agent keeps both tokens per step and total steps low, achieving fast and effective task completion (Figure \ref{fig:inefficiency}(c)).

Existing reinforcement learning (RL) methods serve as a key tool for aligning LLMs with human preferences, with PPO as a representative approach~\citep{schulman2017proximal,luo2025o1,shen2025dast,aggarwal2025l1}. Recent advances further improve training stability (e.g.,  DPO~\citep{rafailov2023direct,chen2025do};  GRPO~\citep{shao2024deepseekmath}) and reduce computational costs (e.g., KTO~\citep{ethayarajh2024model}).
However, these methods focus primarily on learning dynamics rather than agent efficiency—that is, how effectively the agent interacts with the environment. To fill this gap, we propose \sys, a \texttt{D}ual-\texttt{E}fficiency \texttt{P}reference \texttt{O}ptimization method that explicitly targets both step-level and trajectory-level efficiency in LLM agents.
 Concretely, we extend KTO with an efficiency bonus—a parameter‑independent offset added to the desirable log‑ratio—rewarding samples that use fewer steps and tokens. Then we apply the sigmoid preference loss with the Kullback–Leibler (KL) divergence \citep{kullback1951information}. The bonus encourages a larger decision margin for efficient trajectories, amplifying their preference score while leaving the loss formulation and the undesirable branch unchanged.
Overall, \sys relies solely on offline desirable and undesirable labels, without the need for paired annotations, reward model training, or on-policy sampling. The method is stable to train and both compute- and data-efficient.

We conduct comprehensive experiments using multiple LLM agents on five widely adopted benchmarks, Webshop, BabyAI, GSM8K , MATH, and SimulEq. The results show that \sys significantly enhances agent efficiency, achieving up to a 60.9\% reduction in token usage compared to models trained with behavioral cloning (BC), and reducing step count by up to 26.9\% over vanilla KTO. 
Notably, \sys enhances efficiency without sacrificing performance, and even achieves additional performance gains—the largest observed improvement being 29.3\% relative to BC. 
\sys also demonstrates strong generalizability, delivering efficiency improvements on three out-of-domain math benchmarks. Finally, \sys exhibits excellent sample efficiency, maintaining efficiency gains even when trained with only 25\% of the training data.

To summarize, our main contributions are as follows: 
\begin{itemize}
    \item We formally define dual-efficiency for LLM agents in terms of both step-level and trajectory-level efficiency, providing a foundation for developing effective algorithms and evaluating their performance.
    \item We introduce \sys, which incorporates dual-efficiency preferences into vanilla KTO, guiding models to generate responses that use fewer tokens and require fewer steps.
    \item We conduct comprehensive experiments to validate the effectiveness of \sys, and demonstrate its strong generalizability and sample efficiency.
\end{itemize}

% =================== [ SECTION: Related Work ] ================= %

\section{Related Work}
\label{sec:related_work}

\noindent\textbf{LLM agent.}
As LLMs continue to improve in reasoning and planning capabilities, recent research begins to explore the potential of LLMs as agents and investigate ways to further enhance their abilities. Existing approaches can be broadly categorized into three groups: SFT, offline RL, and online RL. 
For SFT-based methods,
AgentTuning~\citep{zeng-etal-2024-agenttuning} uses a trajectory dataset and general-domain instructions to boost LLMs’ agent abilities without harming their general performance. 
Agent-FLAN~\citep{chen-etal-2024-agent} fine-tunes LLMs by decomposing and redesigning the agent training corpus.
AgentRefine~\citep{fu2025agentrefine} uses a data-generation pipeline that simulates diverse environments to overcome overfitting in existing agent-tuning methods. 
For offline RL, ETO~\citep{song-etal-2024-trial} features an exploration–training loop: agents collect their failure trajectories, generate contrastive preference pairs for DPO training. 
DMPO~\citep{shi-etal-2024-direct} adapts DPO for multi-turn agent tasks by using a state-action occupancy measure constraint and adding length normalization to the Bradley–Terry model.
\citet{xiong-etal-2024-watch} introduce the IPR framework, which employs Monte Carlo–estimated step-level rewards to construct contrastive action pairs for DPO training.
As for online RL, \citet{wang2025ragen} propose StarPO, a general reinforcement learning framework designed to optimize the full multi-turn interaction trajectories of LLM agents. \citet{feng2025group} design GiGPO, an RL algorithm that uses a two‐level grouping approach to estimate relative advantage.
However, while these methods primarily focus on improving the performance of LLM agents, research on enhancing their efficiency remains largely unexplored.

\noindent\textbf{Efficient reasoning.}
The phenomenon of overthinking in LLMs has sparked interest in investigating their efficient reasoning.
In addition to controlling inference \citep{pan-etal-2024-dynathink,aytes2025sketch,wang2025make,li2025testtime,han-etal-2025-token} or constructing datasets for supervised fine-tuning \citep{liu2024can,yu2024distilling,kang2025c3ot,xia2025tokenskip,munkhbat-etal-2025-self}, some studies address this problem using RL. The central idea of RL-based approaches is to incorporate a length penalty into the loss function, thereby training LLMs to produce more concise responses \citep{luo2025o1,team2025kimi,shen2025dast,aggarwal2025l1,arora2025training}.
Most of these approaches focus on reducing the tokens generated in a single response, without addressing the inefficiency caused by redundant steps during dynamic interactions between LLM agents and environments.

\section{Method}
\label{sec:method}

\subsection{Setup}

The interaction between an LLM agent and an environment can be modeled as a partially observable Markov decision process (POMDP): ($\mathcal{S},\mathcal{A},\mathcal{O},\mathcal{T},\mathcal{R}$) \citep{song-etal-2024-trial,yuan2025agent}. Here, $\mathcal{S}$ the state space; $\mathcal{A}$ the action space; $\mathcal{O}$ the observation space; the transition function $\mathcal{T}:\mathcal{S}\times\mathcal{A}\to \mathcal{S}$; and the reward function $\mathcal{R}:\mathcal{S}\times\mathcal{A}\to [0,1]$.
Given a task $u$ and an LLM agent  $\pi_\theta$  parameterized by $\theta$, the historical trajectory $\tau_t$ and the action $a_t$ at time $t$ are defined as:
\begin{align}
   \tau_t := (o_1,a_1,\ldots,o_t),\enspace
a_t \sim \pi_\theta(\cdot \mid u,o_{1:t},a_{1:t-1}).
\end{align}
The task terminates upon success or after exceeding the maximum number of rounds, at which point the environment returns a reward $r(\tau)\in[0,1]$.

In this paper, our goal is to train an agent  $\pi_\theta$ that maximizes the expected reward $r(\cdot)$, which involves two objectives: (i) achieving high success rates across tasks, (ii) promoting \textbf{dual efficiency}, which comprises two desiderata — minimizing the number of tokens generated per step and the total number of steps required to complete a task. Efficiency is optimized only among successful trajectories, ensuring necessary long reasoning on difficult tasks remains intact.

To this end, we first generate a large set of trajectories using Monte Carlo Tree Search, and label each as desirable or undesirable based on task success and dual efficiency~(Sec.~\ref{sec:data_generation}). We then adopt a two-stage training procedure: supervised fine-tuning (SFT) on high-quality desirable trajectories~(Sec.~\ref{sec:behavioral_cloning}), followed by our dual-efficiency preference optimization (\sys), which contrasts desirable and undesirable samples to improve generalization under distribution shifts~(Sec.~\ref{sec:depot}).

% ~~~~~~~~~~ [ Subsection: Data Generation ] ~~~~~~~~~~ %

\subsection{Data Generation}
\label{sec:data_generation}

To prepare training data, we follow a two-step process. First, we employ Monte Carlo Tree Search (MCTS) to generate a large set of trajectories. Then, we assign each trajectory as desirable or undesirable based on the corresponding reward.

\paragraph{Step 1: MCTS.}
Inspired by \citet{wang-etal-2024-math,fu2025agentrefine,zhang-etal-2025-arise}, we employ MCTS to generate ReAct-style data \citep{yao2023react}. Each step consists of a natural-language \texttt{Thought} that reasons over the observation, followed by a corresponding \texttt{Action}.
MCTS consists of the following four components:

\begin{itemize}
    \item[1)] \textit{Selection}. 
Starting from the root node, iteratively select child nodes along the trajectory using the upper confidence
bounds (UCT) criterion \citep{kocsis2006bandit}. 
\begin{align}
    \text{UCT}(s)=Q(s)+w\sqrt{\frac{\ln N(\text{Pa}(s))}{N(s)}},
\end{align}
where $Q(s)$ is the average reward of state $s$, $w$ controls the degree of exploration, with higher values encouraging the selection of less frequently visited nodes. $N(\text{Pa}(s))$ refers to the visit count of the parent node of $s$, whereas $N(s)$ represents the visit count of $s$.
\item[2)] \textit{Expansion}. Upon reaching a non-fully expanded node, add a new child node for an available action.
\item[3)] \textit{Rollout}. From the new node, perform a rollout where the model generates actions either until the task is completed or the maximum depth is reached.
\item[4)] \textit{Backpropagation}. Propagate the final reward from the rollout back through the trajectory, updating the values of each node. Through this process, MCTS yields a rich collection of trajectories, which serve as the foundation for the subsequent data filtering and model training stages.
\end{itemize}

\paragraph{Step 2: data  labeling.}
Based on the $r(\tau)$, we design the following protocol to distinguish whether a trajectory is desirable $\tau \in \mathcal{D}$ or undesirable $\tau \in \mathcal{U}$:

\begin{align}
\begin{cases}
\tau \in \mathcal{D},&\ \text{if}\ r(\tau) \geq \kappa_0  \\
\tau \in \mathcal{U},&\ \text{if}\ \kappa_1 > r(\tau) \geq \kappa_2,
\end{cases}
\end{align}
where $1>\kappa_0>\kappa_1>\kappa_2>0$
are  thresholds used to filter trajectories according to $r(\tau)$. A trajectory is labeled as \textbf{desirable} if its reward is greater than or equal to the upper threshold \( \kappa_0 \), indicating strong task performance. Conversely, trajectories with rewards below the lower threshold \( \kappa_2 \) are discarded due to poor quality and limited training value. To facilitate effective preference learning, we retain only \textbf{undesirable} samples whose rewards fall in the intermediate range \( [\kappa_2, \kappa_1) \). By enforcing a margin ($\kappa_0-\kappa_1$) between desirable trajectories (\( r(\tau) > \kappa_0 \)) and undesirable ones (\( \kappa_1 > r(\tau)\)), we ensure a clearer quality separation, making it easier for the model to learn consistent preference signals.

To further improve data quality, we introduce a rephrasing model, which polishes the \texttt{Thought} component at each step using contextual information, while keeping the \texttt{Action} unchanged. Additionally, we ensure that desirable trajectories contain fewer tokens per step after rephrasing. We denote $\mathcal{D}$ as the set of all desirable trajectories and $\mathcal{U}$ the set of all undesirable trajectories. The concrete threshold and rephrasing settings are detailed in Section \ref{sec:setups}.

% ~~~~~~~~~~ [ Subsection: Behavioral Cloning ] ~~~~~~~~~~ %

\subsection{Behavioral Cloning}
\label{sec:behavioral_cloning}
Our pipeline starts with the standard Behavioral Cloning (BC) method for model fine-tuning, a technique that has been widely utilized in prior work~\citep{zeng-etal-2024-agenttuning,song-etal-2024-trial}. 
Specifically,  we randomly select a subset $\mathcal{D}_{\mathsf{BC}}$ from our desirable set $\mathcal{D}$ to serve as the expert demonstrations. Using this dataset,we train an LLM agent to learn a base policy $\pi_\mathsf{BC}$ by minimizing the loss function:
\begin{align}
    \mathcal{L}_{\mathsf{SFT}}(\theta) = -\mathbb{E}_{\tau \sim \mathcal{D}_{\mathsf{BC}}}[ \log \pi_{\theta}(\tau \mid u)].
\end{align}
The expectation is taken with respect to the empirical distribution of trajectories in the dataset $\mathcal{D}_{\mathsf{BC}}$.

% ~~~~~~~~~~ [ Subsection: DEPO ] ~~~~~~~~~~ %
\subsection{Dual-Efficiency Preference Optimization}
\label{sec:depot}
Behavior cloning enables an agent to mimic expert behavior by directly imitating demonstrated thoughts and actions. However, this approach lacks negative signals, which results in brittleness under distribution shift. As outlined in the Section \ref{sec:introduction}, our objective is to improve the agent's \emph{dual efficiency}: (1) minimizing the average number of tokens generated per step, and (2) reducing the total number of steps required to complete a task. To enhance both the efficiency and performance of the agent, we propose to post-train the agent with \sys, an efficiency-aware extension of vanilla KTO that rewards concise thoughts and fewer steps. We start with a brief overview of vanilla KTO.

\noindent\textbf{Vanilla KTO.} Vanilla KTO builds on prospect theory \citep{tversky1992advances}, incorporating reference dependence, diminishing sensitivity, and loss aversion to steer policies toward human-preferred behavior. Formally, KTO has the following optimization objective:
\begin{align}
\mathcal{L}_{\mathsf{KTO}}(\theta) = \mathbb{E}_{\tau \sim \mathcal{D,U}} \left[ \lambda(\tau) - v(\tau) \right],
\end{align}
where the weighting function $\lambda(\tau)$ is defined as
\begin{align}
\lambda(\tau) =
\begin{cases}
\lambda_D, & \text{if } \tau\in \mathcal{D}, \\
\lambda_U, & \text{if } \tau\in \mathcal{U},
\end{cases}
\end{align}
where $\lambda_D, \lambda_U$ are hyper-parameters reflecting the importance of desirable versus undesirable trajectories.
The value function $v(\tau)$ measures the desirability of a trajectory under the current policy, which consists of a implied reward $r_\theta(\tau)$ and a regularization term $z_0(\tau)$:
\begin{align}
v(\tau) = \begin{cases}
\lambda(\tau) \cdot \sigma \left( \beta \left( r_{\theta}(\tau) - z_0(\tau) \right) \right),  \text{if } \tau \in \mathcal{D}, \\
\lambda(\tau) \cdot \sigma \left( \beta \left( z_0(\tau) - r_{\theta}(\tau) \right) \right),  \text{if } \tau \in \mathcal{U},
\end{cases}
\end{align}
where $\sigma(\cdot)$ is the sigmoid function, and $\beta$ is a temperature parameter controlling the sharpness of the sigmoid. The regularization term $z_0(\tau)$ is the KL divergence between the current policy and the baseline policy over the trajectory:
\begin{align}
z_0(\tau) = \mathrm{KL} \left( \pi_{\theta}(\cdot \mid \tau_t) \Vert \pi_{\mathsf{BC}}(\cdot \mid \tau_t) \right).
\end{align}

\noindent\textbf{\sys.} Our \sys~extends vanilla KTO by incorporating an efficiency-aware bonus directly into the reward function $r_\theta(\cdot)$, encouraging trajectories that require fewer tokens per step and fewer steps overall.
Specifically, our implied reward $r_\theta(\tau)$ combines (i) how much the current policy $\pi_\theta$ prefers the observed action, relative to a baseline policy $\pi_{\mathsf{BC}}$, given the trajectory $\tau_t$, and (ii) an efficiency-aware bonus $b(\tau)$ that explicitly favors trajectories with fewer steps and tokens:
\begin{align}
r_{\theta}(\tau) = \log \frac{\pi_{\theta}(a_t \mid \tau_t)}{\pi_{\mathsf{BC}}(a_t \mid \tau_t)} + b(\tau),
\end{align}
where $b(\tau)$ is defined as:
\begin{align}
b(\tau) = 
\begin{cases} 
\frac{\alpha_1}{\overline{T}_{\mathsf{token}}(\tau)} + \frac{\alpha_2}{T_{\mathsf{step}}(\tau)}, 
&\text{if } \tau \in \mathcal{D},
\\
0, &\text{if } \tau\in \mathcal{U},
\end{cases}
\label{eq:depot}
\end{align}
where $\overline{T}_{\mathsf{token}}(\tau)$ denotes the average number of tokens per step in $\tau$, $T_{\mathsf{step}}(\tau)$ is the total number of steps, and $\alpha_1, \alpha_2 > 0$ are hyper-parameters. 
Larger values of \( \overline{T}_{\mathsf{token}}(\tau) \) or \( T_{\mathsf{step}}(\tau) \) reduce the efficiency bonus \( b(\tau) \), thereby lowering the overall reward \( r_\theta(\tau) \) and discouraging inefficient trajectories during preference optimization.

Note that the bonus term \( b(\tau) \) is applied only to desirable trajectories \( \tau \in \mathcal{D} \), and set to zero for undesirable ones \( \tau \in \mathcal{U} \). This selective design simplifies implementation and avoids injecting unnecessary signal into low-quality samples.  
Such design is also supported by our empirical results (Section~\ref{sec:ablation_studies}), which show that penalizing undesirable trajectories offers no additional performance benefit.

% =================== [ SECTION: Experiment ] ================= %

\section{Experiment}
\label{sec:experiment}

% ~~~~~~~~~~ [ Subsection: Setups ] ~~~~~~~~~~ %

\begin{table*}[t!]
\centering
\small
\renewcommand{\arraystretch}{1.2} 
\setlength{\tabcolsep}{1.5pt} 
\fontsize{9}{10}\selectfont
% \setlength{\tabcolsep}{8pt} 
% \scalebox{0.85}
{
\begin{tabular}{l| >{\centering\arraybackslash}p{1.1cm} >{\centering\arraybackslash}p{1.1cm}>{\centering\arraybackslash}p{1.1cm}>{\centering\arraybackslash}p{1.1cm}
                                >{\centering\arraybackslash}p{1.2cm} >{\centering\arraybackslash}p{1.1cm}|
                                >{\centering\arraybackslash}p{1.1cm} >{\centering\arraybackslash}p{1.1cm}>{\centering\arraybackslash}p{1.1cm}>{\centering\arraybackslash}p{1.1cm}
                                >{\centering\arraybackslash}p{1.2cm} >{\centering\arraybackslash}p{1.1cm}}
\toprule
\multirow{2}{*}{\textbf{Method}} &  \multicolumn{6}{c|}{\textbf{Webshop}} &  \multicolumn{6}{c}{\textbf{BabyAI}}  \\ 

                        & \textbf{Succ.}  & \textbf{Re.} &\textbf{T@All} &\textbf{S@All}  & \textbf{T@Succ.}  & \textbf{S@Succ.}  & \textbf{Succ.}  &  \textbf{Re.} & \textbf{T@All} &\textbf{S@All}  & \textbf{T@Succ.}  & \textbf{S@Succ.}  \\
&(↑)&(↑)&(↓)&(↓)&(↓)&(↓) &(↑)&(↑)&(↓)&(↓)&(↓)&(↓)\\
\midrule
DeepSeek-V3& 0.11&  0.27& 6769
&36.19& 439
& 6.8& 0.61&  0.57& 2235
&24.53& 489
&7.69
\\ 
R1-Distill-Qwen-7B& 0.00&  0.00& 6639
&3.09& 0
& 0.00& 0.09&  0.08& 5719
&15.30& 317
& 3.75
\\  
R1-Distill-Llama-8B& 0.00&  0.00& 7360
&5.32& 0
& 0.00& 0.02&  0.02& 6047
&19.60& 784
& 6.00
\\  
Llama-3.1-8B-Instruct& 0.03&  0.31& 501
&13.57& 232
& 8.00& 0.78&  0.74& 604
&16.89& 304
& 8.13
\\  
Llama-3.2-3B-Instruct& 0.00&  0.00& 831
&17.06& 0
& 0.00& 0.68&  0.62& 571
&19.89& 219
& 7.56
\\  
Qwen2.5-3B-Instruct& 0.00&  0.02& 1483
&18.80& 0
& 0.00& 0.40&  0.37& 2187
&23.78& 487
& 7.92
\\  
Qwen2.5-7B-Instruct& 0.02&  0.13& 2094
&17.82& 372
& 5.50& 0.57&  0.52& 1271
&20.09& 431
& 7.49
\\
Qwen2.5-14B-Instruct& 0.09&  0.19& 990
&12.89& 319
& 6.25& 0.64&  0.51& 1026
&19.26& 449
& 8.33
\\  
Qwen2.5-72B-Instruct& 0.06&  0.08& 1622
&17.51& 361
& 8.80& 0.39&  0.36& 2295
&26.72& 667
& 8.57
\\
\midrule
Llama-3.1-8B-BC           & 0.47&  \textbf{0.79}& 840
&6.38& 812
& 6.50& 0.77&  0.71& 836
&13.09& 365
& 5.61
\\  
\quad + TB& 0.42&  0.76& 791
& \textbf{6.32}& 793
& \textbf{6.34}& 0.82&  0.70& 783& 12.21& 399& 6.30\\  
\quad + KTO& 0.48&  0.67& 776
&8.96& {567}
& 6.88& 0.87&  0.81& 342
&10.14& \textbf{145}
& 4.35
\\  
\quad + \textbf{\sys}& \textbf{0.50}&  0.72& \textbf{633}
&7.80& \textbf{500}
& 6.83& \textbf{0.88}&  \textbf{0.82}& \textbf{327}
&\textbf{9.32}& \textbf{145}
& \textbf{4.09}
\\  
\midrule
Qwen2.5-7B-BC           & 0.44&  0.75& 1014
&\textbf{6.34}& 828
&\textbf{6.25}& 0.47&  0.43& 2062
&22.51& 641
& 5.98
\\ 
\quad + TB& 0.42&  0.73& 803
& 6.52& 803
& {6.29}& 0.59&  0.53& 1780
&19.60& 589
& 6.96
\\  
\quad + KTO& 0.54&  0.76& 886
&8.18& \textbf{645}
& 6.88& 0.58&  0.57& 1199
&22.81& \textbf{350}
& \textbf{5.67}
\\  
\quad + \textbf{\sys}& \textbf{0.56}&  \textbf{0.80}& \textbf{726}&7.73& 669& 7.35& \textbf{0.75}&  \textbf{0.69}& \textbf{893}&\textbf{16.67}& 464& 7.03\\  
\bottomrule
\end{tabular}}
\caption{Comparison of \sys with a wide range of baselines. ``-BC'' denotes models cold-started via behavioral cloning. Metrics use the following abbreviations: Succ. (success rate), Re. (mean trajectory reward), T@All (mean tokens per trajectory over all trajectories), S@All (mean steps per trajectory over all trajectories), and T@Succ., S@Succ. (computed only over successful trajectories). Higher is better for Succ. and Reward; lower is better for T and S. ``R1-Distill'' denotes DeepSeek-R1-Distill models. Models with Succ. below 0.2 are excluded from the final comparison, as assessing efficiency under such low Succ. is not meaningful. Best results in each group are in bold.}
\label{tab:main}
\end{table*}

\subsection{Setup}
\label{sec:setups}

\noindent\textbf{Baselines.}
We consider a wide range of baselines, including Deepseek-V3 \citep{liu2024deepseek}, DeepSeek-R1-Distill-\{Qwen-7B, Llama-8B\} \citep{guo2025deepseek}, Llama-3.1-8B-Instruct \citep{dubey2024llama}, Llama-3.2-3B-Instruct \citep{meta2024llama}, Qwen2.5-\{3B,7B,72B\}-Instruct \citep{qwen2025qwen25technicalreport}. In terms of training paradigms, we also compare BC and vanilla KTO. Additionally, we include Token Budget (TB) \citep{han-etal-2025-token}, which compresses output length by including a reasonable budget directly in the prompt.

\noindent\textbf{Datasets.}
Following previous work \citep{xi-etal-2025-agentgym,fu2025agentrefine,yuan2025agent}, we evaluate our \sys on five diverse datasets: Webshop \citep{yao2022webshop}, BabyAI \citep{chevalier-boisvert2018babyai}, GSM8K \citep{cobbe2021training}, MATH \citep{hendrycks2021measuring}, and SimulEq \citep{kushman-etal-2014-learning}. Webshop is an interactive environment that simulates online shopping, where the agent must follow natural language instructions and perform actions such as clicking or searching to locate and purchase the desired item. In contrast, BabyAI is a classic grid-world environment focused on embodied interaction, designed to assess an agent’s ability to comprehend and carry out basic commands such as ``go to the red ball'' or ``pick up the key''. GSM8K, MATH, and SimulEq are three widely used math benchmarks.

\noindent\textbf{Prompts.}
We use the ReAct-style 0-shot prompt \citep{yao2023react}, directing the model to respond in the ``\texttt{Thought}: ...; \texttt{Action}: ...'' format. We do not rely on heavy prompt engineering such as ICL \citep{brown2020language} or CoT \citep{wei2022chain}, nor do we provide overly detailed instructions. This design simulates real-world usage and better assesses the model's generalization ability when user prompts are not meticulously crafted.

\noindent\textbf{Metrics.}
We evaluate the agents from two perspectives: efficiency and performance. For efficiency, we consider token count and step count. Token count is measured  with the \texttt{cl100k\_base} encoding from tiktoken \citep{openai2025tiktoken}, while step count refers to the number of responses the model generates to complete an entire trajectory. 
For performance, we report both success rate and reward. For Webshop and BabyAI, we define task success based on environment feedback on the final step. The reward measures the agent’s average reward obtained across episodes \citep{xi-etal-2025-agentgym}.

\noindent\textbf{Implementation details.}
For MCTS, we set the maximum search depth to 50 and employ Deepseek-V3 for the search process.
For data filtering, in BabyAI, trajectories with $0.9 \leq r(\tau)$ are assigned to the desirable set $D$, while those with $0.7 \leq r(\tau) < 0.9$ are included in the undesirable set $U$. In Webshop, trajectories with $r(\tau) = 1$ are considered desirable, and the criteria for undesirable trajectories follow the same range as in BabyAI.
Next, we filter the data based on the number of steps: trajectories with steps $< 7$ are included in set $\mathcal{D}$, while those with steps $\geq 7$ are assigned to set $\mathcal{U}$. Finally, rephrasing is performed using GPT-4.1 mini \citep{openai2025gpt4}  with a temperature setting of 0.7.
For BC, we use 972 trajectories from BabyAI and 3,732 from Webshop. Fine-tuning is performed using LoRA \citep{hu2022lora} with a learning rate of 1.0e-4 for 3 epochs. For dual-efficiency preference optimization, the BabyAI contains 512 desirable and 471 undesirable trajectories, while Webshop provides 1,567 for each category. We set $\beta$ to 0.2, $\lambda_D$ and $\lambda_U$ to 1, with a learning rate of 2.0e-5 and 3 epochs. For \bcllama, both $\alpha_1$ and $\alpha_2$ are set to 3; for \bcqwen, both are set to 2. All experiments are conducted with eight NVIDIA Tesla A800 GPUs, each with 80GB of memory.

\begin{figure*}[t!]
\includegraphics[width=\textwidth]{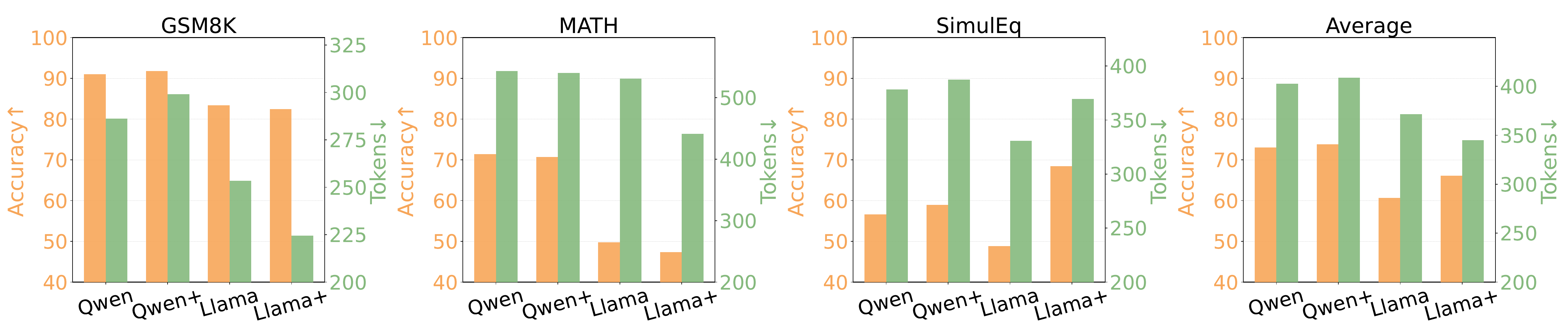}
  \caption{Model generalizability across math benchmarks. The left y‑axis shows accuracy and the right y‑axis shows average tokens. Qwen refers to \qwen, Qwen+ to \bcqwen+\sys, Llama to \llama, Llama+ to \bcllama+\sys.}
\label{fig:generalizability}
\end{figure*}

% ~~~~~~~~~~ [ Subsection: Main Results ] ~~~~~~~~~~ %

\subsection{Main Results}
\label{sec:main_results}

Table \ref{tab:main} presents a detailed comparison between \sys and various baselines. We draw the following conclusions:

\noindent\textbf{\sys achieves a significant improvement in the efficiency of LLM agents.}
In terms of token count, \sys achieves the lowest token usage across both the Webshop and BabyAI datasets. Compared to vanilla KTO, \bcllama+\sys reduces T@All, T@Succ. on Webshop and T@All on BabyAI by $18.4\%$, $11.8\%$, and $4.4\%$ respectively. When compared to \bcllama, the reductions in T@All and T@Succ. on both datasets are even more pronounced, with decreases of $24.6\%$, $38.4\%$, $60.9\%$, and $60.3\%$.
\bcqwen+\sys achieves reductions of $18.1\%$ and $25.5\%$ in T@All on Webshop and BabyAI compared to vanilla KTO. When compared to \bcqwen, T@All and T@Succ. decrease by $28.4\%$, $19.2\%$, $56.7\%$, and $27.6\%$ across the two datasets.
Regarding step count, compared to vanilla KTO, \bcllama+\sys reduces S@All and S@Succ. by 12.9\% and 0.7\% on Webshop, and by 8.1\% and 6.0\% on BabyAI. As for \bcqwen+\sys, S@All decreases by 5.5\% on Webshop and by 26.9\% on BabyAI.

\noindent\textbf{\sys not only ensures efficiency but also maintains—or even improves—the performance of LLM agents.}
Compared with KTO and TB, \sys offers a better performance/efficiency trade‑off: it reduces T@All/T@Succ. and S@All/S@Succ. by similar or larger margins while delivering higher Succ. and Reward.
\bcqwen+\sys achieves the highest Succ. and Reward on Webshop, while \bcllama+\sys attains the best results on BabyAI. Additionally, \bcllama+\sys improves Succ. and Reward on Webshop by 4.2\% and 7.5\% over vanilla KTO, whereas \bcqwen+\sys boosts these metrics on BabyAI by 29.3\% and 21.1\%.

\noindent\textbf{The primary advantage of \sys on Webshop lies in its substantial reduction of tokens per step.}
This trend holds for both \bcllama+\sys and \bcqwen+\sys: although the number of steps increases slightly, the average tokens per step (i.e., T@All/S@All and T@Succ./S@Succ.) decrease significantly, leading to a notable reduction in total tokens. 
For \bcllama+\sys, T@All/S@All drops by 6.4\% and T@Succ./S@Succ. by 11.2\%. For \bcqwen+\sys, the reductions are 13.3\% and 2.8\%. All values are relative to vanilla KTO.
% 

% ~~~~~~~~~~ [ Subsection: Generalizability ] ~~~~~~~~~~ %

\subsection{Generalizability}
\label{sec:generalizability}

To evaluate the generalizability and performance retention of \sys, we compare models trained with \sys to their original Instruct versions on GSM8K, MATH, and SimulEq. Figure \ref{fig:generalizability} presents the results, showing Accuracy (left axis, higher is better) and Tokens (right axis, lower is better) for each model.
We find that: 

\noindent \textbf{\sys demonstrates strong generalizability, achieving higher average accuracy while reducing generation tokens on most tasks.}
\sys consistently improves the accuracy of both models on SimulEq as well as on the average across all three datasets. In terms of efficiency, \bcllama+\sys achieves a clear reduction in token usage, while \bcqwen+\sys shows a slight increase on average; for the MATH dataset, both models generally generate fewer tokens.

% ~~~~~~~~~~ [ Subsection: Sample Efficiency ] ~~~~~~~~~~ %
\subsection{Sample Efficiency}
\label{sec:sample_efficiency}

\begin{figure}[t!]  
\centering
\includegraphics[width=.85\columnwidth]{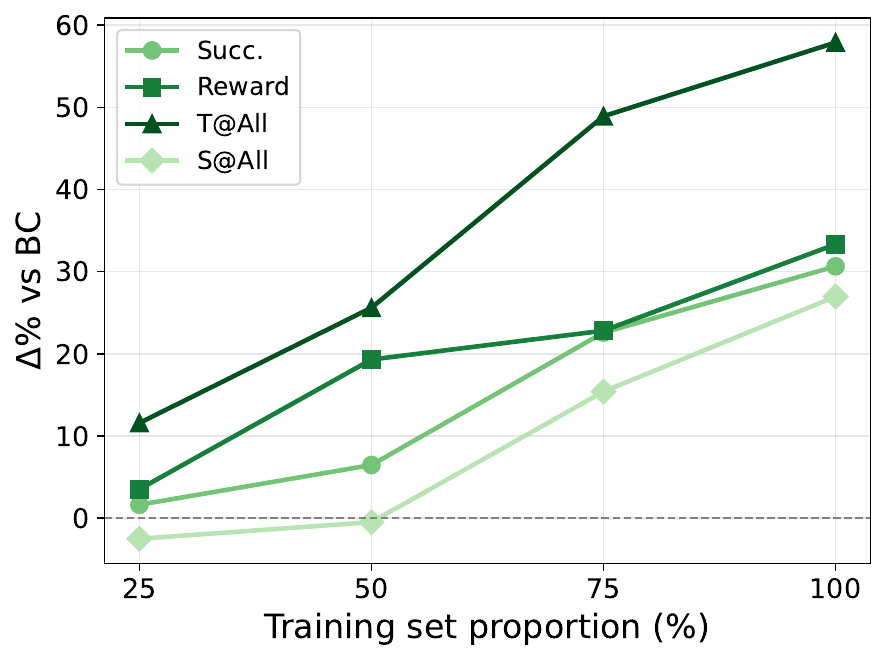}
  \caption{Sample efficiency of \sys. Relative improvement $\triangle$\% over the BC baseline for four metrics as the training‐set proportion increases from 25\% to 100\%.}
\label{fig:validity}
\end{figure}

We evaluate the sample efficiency of our method by training with 25\%, 50\%, 75\%, and 100\% of the full dataset, and compare both agent's performance and efficiency across these settings. We report the average results of \bcllama~ and \bcqwen~on BabyAI, as shown in Figure \ref{fig:validity}. 
Our key findings are as follows:

\noindent \textbf{\sys exhibits excellent sample efficiency.} With just 25\% of the training data (only 245 samples for BabyAI and 783 for Webshop), it  delivers substantial improvements over the BC baseline. In particular, T@All increases efficiency by more than 10\%. This demonstrates that \sys is able to rapidly acquire preferences and boost both performance and efficiency, even when training data is highly limited.
    
\noindent \textbf{With larger training sets, \sys consistently yields further gains.} All four metrics improve as the amount of data increases, and T@All, in particular, achieves nearly a 60\% improvement with the full dataset. These results suggest that \sys is able to effectively utilize more data for continued improvement, and is less prone to early saturation.

% ~~~~~~~~~~ [ Subsection: Ablation Studies ] ~~~~~~~~~~ %

\subsection{Ablation Studies}
\label{sec:ablation_studies}

\paragraph{Effect of hyper-parameters $\alpha_1$ and $\alpha_2$.}

\begin{table*}[t!]
\centering
\small
\renewcommand{\arraystretch}{1.2} 
\fontsize{9}{10}\selectfont
\setlength{\tabcolsep}{1.5pt} 
% \scalebox{0.85}
{
\begin{tabular}{c| >{\centering\arraybackslash}p{1.1cm} >{\centering\arraybackslash}p{1.1cm}>{\centering\arraybackslash}p{1.1cm}>{\centering\arraybackslash}p{1.1cm}
                                >{\centering\arraybackslash}p{1.2cm} >{\centering\arraybackslash}p{1.1cm}|
                                >{\centering\arraybackslash}p{1.1cm} >{\centering\arraybackslash}p{1.1cm}>{\centering\arraybackslash}p{1.1cm}>{\centering\arraybackslash}p{1.1cm}
                                >{\centering\arraybackslash}p{1.2cm} >{\centering\arraybackslash}p{1.1cm}}
\toprule
\multirow{2}{*}{\textbf{Parameter}} &  \multicolumn{6}{c|}{\textbf{Webshop}} &  \multicolumn{6}{c}{\textbf{BabyAI}}  \\ 

                        & \textbf{Succ.}  & \textbf{Re.} &\textbf{T@All} &\textbf{S@All}  & \textbf{T@Succ.}  & \textbf{S@Succ.}  & \textbf{Succ.}  &  \textbf{Re.} & \textbf{T@All} &\textbf{S@All}  & \textbf{T@Succ.}  & \textbf{S@Succ.}  \\
&(↑)&(↑)&(↓)&(↓)&(↓)&(↓) &(↑)&(↑)&(↓)&(↓)&(↓)&(↓)\\
\midrule
\multicolumn{13}{c}{\bcllama} \\
\midrule
$\alpha_1=3,\alpha_2=3$& \textbf{0.50}&  \textbf{0.72}& \textbf{633}
&\textbf{7.80}& 500
& 6.83
& \textbf{0.88}&  \textbf{0.82}& 327
&\textbf{9.32}& \textbf{145}
&\textbf{4.09}
\\ 
$\alpha_1=3,\alpha_2=0$& 0.43&  0.66& 818
&8.86& \textbf{459}
& \textbf{6.51}
& \textbf{0.88}&  \textbf{0.82}& \textbf{326}
&9.71& 160
& 4.47
\\  
$\alpha_1=0,\alpha_2=3$& 0.43&  0.67& 849&9.27& 549& 6.62& 0.84&  0.78& 458&12.48& 233& 6.01\\  
\midrule
\multicolumn{13}{c}{\bcqwen} \\
\midrule
$\alpha_1=2,\alpha_2=2$ & \textbf{0.56}&  \textbf{0.80}& 726
&\textbf{7.73}& 669
& 7.35
& \textbf{0.75}&  \textbf{0.69}& \textbf{893}
&\textbf{16.67}& 464
& 7.03
\\  
$\alpha_1=2,\alpha_2=0$ & 0.49&  0.72& 757
& 8.00& 670
& 7.30
& 0.66&  0.61& 1149
& 19.68& 410
& \textbf{5.81}
\\  
$\alpha_1=0,\alpha_2=2$ & 0.51&  0.74& \textbf{703}&7.98& \textbf{657}& \textbf{7.00}& 0.63&  0.59& 1059&21.26& \textbf{391}& 6.58\\   
\bottomrule
\end{tabular}}
\caption{The impact of different $\alpha_1$, $\alpha_2$ values on model performance and efficiency. The best result for each model is highlighted in bold, with comparisons made within each model group rather than across models. The results show that jointly considering both $\alpha_1$ and $\alpha_2$ achieves the best trade-off between performance and efficiency.
}
\label{tab:alpha}
\end{table*}
As defined by Eq.~\eqref{eq:depot}, $\alpha_1$ and $\alpha_2$ are parameters that adjust step-level and trajectory-level efficiency, respectively. In Table \ref{tab:alpha}, we compare the impact of different values of these parameters on both performance and efficiency.
We observe that: 

\noindent\textbf{Jointly considering both $\alpha_1$ and $\alpha_2$ achieves the best overall trade-off between performance and efficiency.}
For \bcllama, setting $\alpha_1 = 3$ and $\alpha_2 = 3$ on Webshop yields the highest Succ. and Reward, along with the beat T@All and S@All. On BabyAI, this configuration matches the performance of $\alpha_1 = 3$, $\alpha_2 = 0$ but demonstrates superior efficiency across multiple metrics.
For \bcqwen, $\alpha_1 = 2$ and $\alpha_2 = 2$ achieves the best Succ., Reward, and S@All on Webshop, with T@All being the second best. On BabyAI, it attains the highest scores across Succ., Reward, T@All, and S@All.

\noindent \textbf{Optimizing a single parameter can further improve individual efficiency metrics, but this often comes at the expense of performance.}
When emphasizing step-level efficiency ($\alpha_1>0, \alpha_2=0$), \bcllama~achieves the lowest T@Succ. on Webshop and the lowest T@All on BabyAI. However, its performance is only comparable to, or slightly lower than, the balanced setting. Conversely, when focusing on trajectory-level efficiency ($\alpha_1=0, \alpha_2>0$), \bcqwen~attains the lowest S@Succ. on Webshop, but both Succ. and Reward drop noticeably.

\paragraph{Undesirable penalty.}

\begin{figure}[t!]  
\centering
\includegraphics[width=.85\columnwidth]{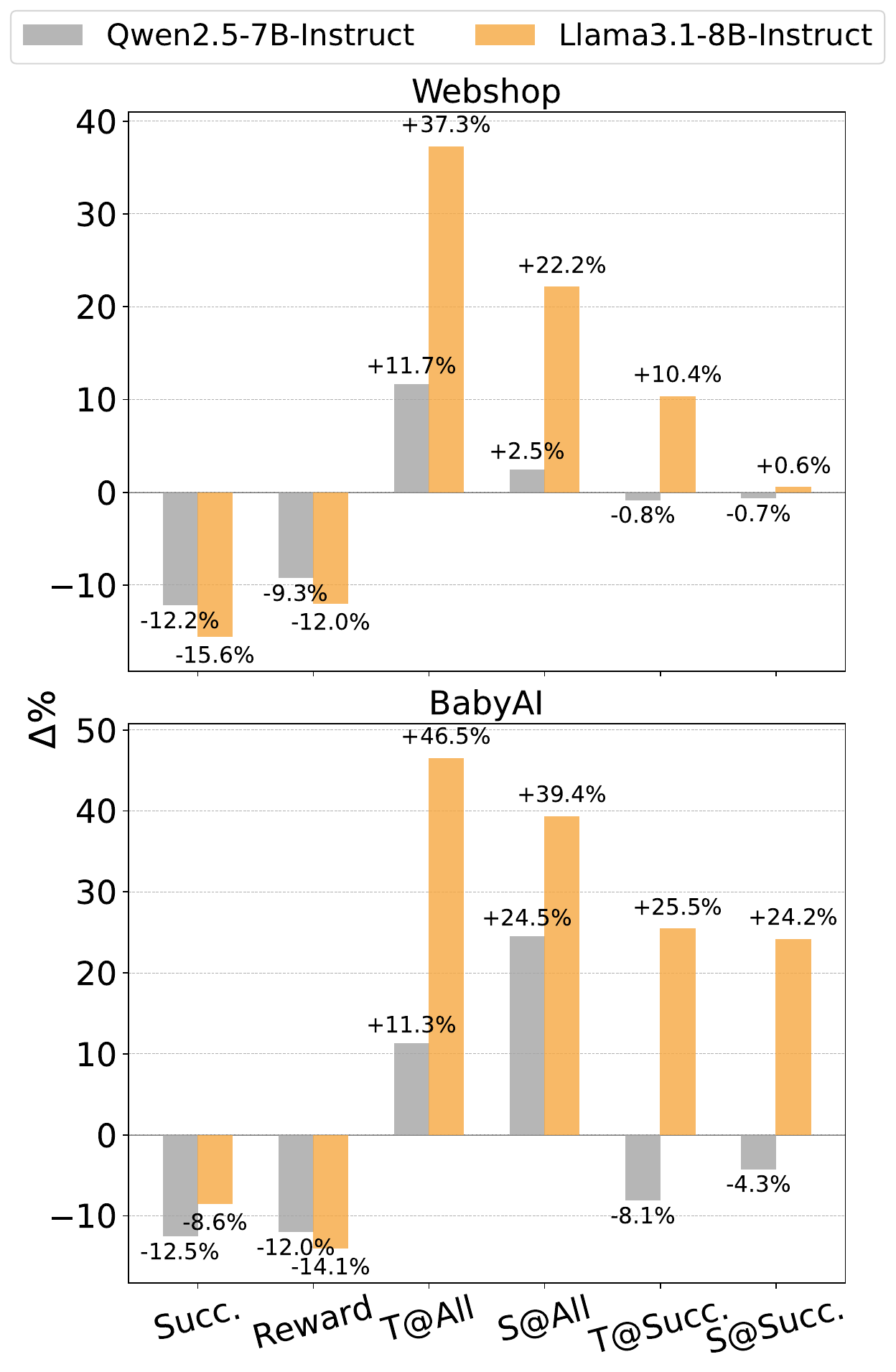}
  \caption{Desirable bonus only (\sys) vs. Desirable bonus (\sys) + Undesirable penalty. Each point shows the relative change of \sys+ undesirable penalty against the \sys, with $\triangle$\% = (Penalty $-$ \sys)/\sys × 100.}
\label{fig:penalty}
\end{figure}

As defined in Section \ref{sec:depot}, we only apply the bonus to desirable samples. A natural extension is to ask whether we can also impose a penalty on undesirable samples. In this way, penalty can be formulated as follows: 
\begin{align}
p(\tau) = 
\begin{cases} 
\frac{\alpha_1}{\overline{T}_{\text{token}}(\tau)} + \frac{\alpha_2}{T_{\text{step}}(\tau)}, 
&\text{if } \tau \in \mathcal{U},
\\
0, &\text{if } \tau\in \mathcal{D},
\end{cases}
\label{eq:penalty}
\end{align}
We can then similarly add the penalty offset back to the undesirable trajectory.
In this ablation experiment, we apply equally strong bonuses and penalties, with $\alpha_1 = 3, \alpha_2 = 3$ for \bcllama~and $\alpha_1 = 2, \alpha_2 = 2$ for \bcqwen. Figure \ref{fig:penalty} shows a comparison between \sys and \sys+ penalty.

We find that: \textbf{applying an additional penalty to undesirable trajectories with the same intensity does not yield overall improvements.} In terms of performance, both models exhibit noticeable declines on Webshop and BabyAI when the penalty is applied. Regarding efficiency, T@All and S@All generally rise markedly, with the most pronounced increases observed for \bcllama~on BabyAI (T@All +46.5\%, S@All +39.4\%). Although \bcqwen~exhibits slight reductions in T@Succ. and S@Succ., these gains are insufficient to counterbalance the broader negative impact.

% =================== [ SECTION: Conclusion ] ================= %

\section{Conclusion}
\label{sec:conclusion}

In this work, we introduce dual-efficiency for LLM agents, encompassing both step-level efficiency (fewer tokens per turn) and trajectory-level efficiency (fewer steps per task). Based on this definition, we design \sys, a preference-based optimization method that incorporates an efficiency bonus into the desirable log-ratio, rewarding samples that achieve tasks with fewer steps and tokens.
\sys relies solely on offline preference data, without the need for paired annotations, reward model training, or on-policy sampling.
Extensive experiments demonstrate that \sys effectively improves agent efficiency while enhancing task performance. Moreover, \sys exhibits strong generalizability across domains and maintains high sample efficiency. 

\section*{Acknowledgments}
We thank all the anonymous reviewers and area
chair for their valuable feedback throughout the review process. This work is supported in part by Shanghai
Artificial Intelligence Laboratory and in part by the National Key R\&D Program of China 2023YFC3806000. 

\bibliography{aaai2026}

\end{document}